\documentclass[10pt,twocolumn,letterpaper]{article}
\usepackage{iccv}
\usepackage{times}
\usepackage{epsfig}
\usepackage{graphicx}
\usepackage{amsmath}
\usepackage{amssymb}

\usepackage{color}
\usepackage{multirow}
\usepackage{amssymb}
\usepackage{pifont}% http://ctan.org/pkg/pifont
\newcommand{\cmark}{\ding{51}}%
\newcommand{\xmark}{\ding{55}}%
\usepackage[pagebackref=true,breaklinks=true,letterpaper=true,colorlinks,bookmarks=false]{hyperref}

\iccvfinalcopy % *** Uncomment this line for the final submission

\def\iccvPaperID{3649} % *** Enter the ICCV Paper ID here

\ificcvfinal\fi

\begin{document}

%%%%%%%%% TITLE
\title{Channel-wise Topology Refinement Graph Convolution for Skeleton-Based Action Recognition}

\author{Yuxin Chen\textsuperscript{1,2}, Ziqi Zhang\textsuperscript{1,2}, Chunfeng Yuan\textsuperscript{1}\thanks{Corresponding author.}, Bing Li\textsuperscript{1}, Ying Deng\textsuperscript{4}, Weiming Hu\textsuperscript{1,3}\\
\textsuperscript{1}NLPR, Institute of Automation, Chinese Academy of Sciences\\
\textsuperscript{2}School of Artificial Intelligence, University of Chinese Academy of Sciences\\
\textsuperscript{3}CAS Center for Excellence in Brain Science and Intelligence Technology\\
\textsuperscript{4}School of Aeronautical Manufacturing Engineering, Nanchang Hangkong University\\
{\tt\small chenyuxin2019@ia.ac.cn, \{ziqi.zhang,cfyuan,bli,wmhu\}@nlpr.ia.ac.cn}

%\and
%Ying Deng\\
%School of Aeronautical Manufacturing Engineering, Nanchang HangKong University\\
}

\maketitle
\ificcvfinal\thispagestyle{empty}\fi

%%%%%%%%% ABSTRACT
\begin{abstract}
   Graph convolutional networks (GCNs) have been widely used and achieved remarkable results in skeleton-based action recognition. In GCNs, graph topology dominates feature aggregation and therefore is the key to extracting representative features. In this work, we propose a novel Channel-wise Topology Refinement Graph Convolution (CTR-GC) to dynamically learn different topologies and effectively aggregate joint features in different channels for skeleton-based action recognition. The proposed CTR-GC models channel-wise topologies through learning a shared topology as a generic prior for all channels and refining it with channel-specific correlations for each channel. Our refinement method introduces few extra parameters and significantly reduces the difficulty of modeling channel-wise topologies. Furthermore, via reformulating graph convolutions into a unified form, we find that CTR-GC relaxes strict constraints of graph convolutions, leading to stronger representation capability. Combining CTR-GC with temporal modeling modules, we develop a powerful graph convolutional network named CTR-GCN which notably outperforms state-of-the-art methods on the NTU RGB+D, NTU RGB+D 120, and NW-UCLA datasets.\footnote{ \url{https://github.com/Uason-Chen/CTR-GCN}.}
\end{abstract}
%The proposed CTR-GC infers channel-wise topologies by explicitly modeling pairwise correlations between vertices in each channel. In addition, the inference difficulty is notably reduced by performing refinement on a channel-shared topology.
%%%%%%%%% BODY TEXT
\section{Introduction}

Human action recognition is an important task with various applications ranging from human-robot interaction to video surveillance. In recent years, skeleton-based human action recognition has attracted much attention due to the development of depth sensors and its robustness against complicated backgrounds.

\begin{figure}[t]
	\centering
	\includegraphics[width=\columnwidth]{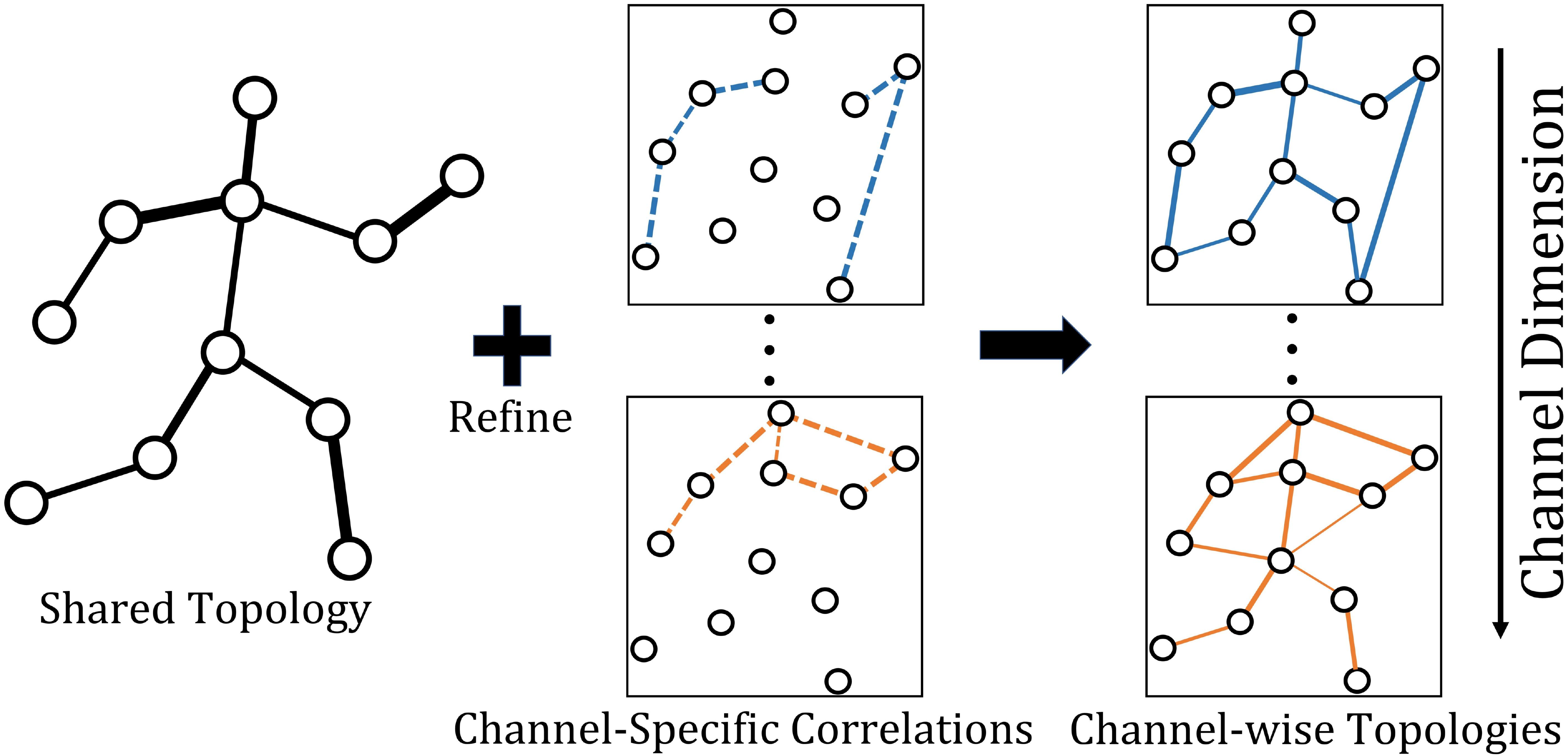} % Reduce the figure size so that it is slightly narrower than the column. Don't use precise values for figure width.This setup will avoid overfull boxes.
	\caption{Channel-wise topology refinement. Lines of different colors correspond to topologies in different channels and the thickness of lines indicates the correlation strength between joints.}
	\label{fig:adaptive_correlation}
	\vspace{-0.4cm}
\end{figure}

Early deep-learning-based methods treat human joints as a set of independent features and organize them into a  feature sequence or a pseudo-image, which is fed into RNNs or CNNs to predict action labels. However, these methods overlook inherent correlations between joints, which reveals human body topology and is important information of human skeleton. Yan \etal \cite{yan2018spatial} firstly modeled correlations between human joints with graphs and apply GCNs along with temporal convolutions to extract motion features. While the manually defined topology they employ is difficult to achieve relationship modeling between unnaturally connected joints and limits representation capability of GCNs. In order to boost power of GCNs, recent approaches \cite{shi2019two,zhang2020semantics,ye2020dynamic} adaptively learn the topology of human skeleton through attention or other mechanisms. They use a topology for all channels, which forces GCNs to aggregate features with the same topology in different channels and thus limits the flexibility of feature extraction. Since different channels represent different types of motion features and correlations between joints under different motion features are not always the same, it's not optimal to use one shared topology. Cheng \etal \cite{cheng2020eccv} set individual parameterized topologies for channel groups. 
However, the topologies of different groups are learned independently and the model becomes too heavy when setting channel-wise parameterized topologies, which increases the difficulty of optimization and hinders effective modeling of channel-wise topologies. Moreover, parameterized topologies remain the same for all samples, which is unable to model sample-dependent correlations.

In this paper, we propose a channel-wise topology refinement graph convolution which models channel-wise topology dynamically and effectively. Instead of learning topologies of different channels independently, CTR-GC learns channel-wise topologies in a refinement way. Specifically, CTR-GC learns a shared topology and channel-specific correlations simultaneously. The shared topology is a parameterized adjacency matrix that serves as topological priors for all channels and provides generic correlations between vertices. The channel-specific correlations are dynamically inferred for each sample and they capture subtle relationships between vertices within each channel. By refining the shared topology with channel-specific correlations, CTR-GC obtains channel-wise topologies (illustrated in Figure \ref{fig:adaptive_correlation}). Our refinement method avoids modeling the topology of each channel independently and introduces few extra parameters, which significantly reduces the difficulty of modeling channel-wise topologies. 
Moreover, through reformulating four categories of graph convolutions into a unified form, we verify the proposed CTR-GC essentially relaxes strict constraints of other categories of graph convolutions and improves the representation capability.

Combining CTR-GC with temporal modeling modules, we construct a powerful graph convolutional network named CTR-GCN for skeleton-based action recognition. Extensive experimental results on NTU RGB+D, NTU RGB+D 120, and NW-UCLA show that (1) our CTR-GC significantly outperforms other graph convolutions proposed for skeleton-based action recognition with comparable parameters and computation cost; (2) Our CTR-GCN exceeds state-of-the-art methods notably on all three datasets.
% (2) Our CTR-GC can effectively learn subtle correlations, especially interactions between fingers;

Our contributions are summarized as follows:

\begin{itemize}
	\item We propose a channel-wise topology refinement graph convolution which dynamically models channel-wise topologies in a refinement approach, leading to flexible and effective correlation modeling.
	\item We mathematically unify the form of existing graph convolutions in skeleton-based action recognition and find that CTR-GC relaxes constraints of other graph convolutions, providing more powerful graph modeling capability.
	\item The extensive experimental results highlight the benefits of channel-wise topology and the refinement method. The proposed CTR-GCN outperforms state-of-the-art methods significantly on three skeleton-based action recognition benchmarks.
\end{itemize}

%-------------------------------------------------------------------------
\section{Related Work}

%In this section, we provide a brief review on GCNs and GCN-based skeleton-based action recognition methods.
\subsection{Graph Convolutional Networks}

Convolutional Neural Networks (CNNs) have achieved remarkable results in processing Euclidean data like images. To process non-Euclidean data like graphs, there is an increasing interest in developing Graph Convolutional Networks (GCNs). GCNs are often categorized as spectral methods and spatial methods. Spectral methods conduct convolution on spectral domain \cite{bruna2013spectral,defferrard2016convolutional,kipf2016semi}. However, they depend on the Laplacian eigenbasis which is related to graph structure and thus can only be applied to graphs with same structure. Spatial methods define convolutions directly on the graph \cite{duvenaud2015convolutional,niepert2016learning,velivckovic2017graph}. One of the challenges of spatial methods is to handle different sized neighborhoods. Among different GCN variants, the GCN proposed by Kipf \etal \cite{kipf2016semi} is widely adapted to various tasks due to its simplicity. The feature update rule in \cite{kipf2016semi} consists of two steps: (1) Transform features into high-level representations; and (2) Aggregate features according to graph topology. Our work adopts the same feature update rule. %, \ie, feature transformation and feature aggregation.
%One of the challenges of spatial methods is to handle different sized neighborhoods while keeping the weight sharing property of CNNs. Duvenaud \etal \cite{duvenaud2015convolutional} proposed to learn individual weight matrix for each node degree. Niepert \etal \cite{niepert2016learning} normalized neighborhoods to contain a fixed number of nodes. Inspired by attention mechanisms, Veli\v{c}kovi\'{c} \etal \cite{velivckovic2017graph} proposed a Graph Attention Network which adopts attention mechanisms to learn the relative correlations between two connected vertices, however, learned correlations are shared among different channels which limit the model capability.
\subsection{GCN-based Skeleton Action Recognition}

%The advanced depth sensors and pose estimation algorithms boost the development of skeleton-based action recognition. Methods on this task can be categorized as RNN-based, CNN-based and GCN-based methods.

%RNN-based methods organize skeleton data into a sequence which is fed into RNNs to predict action label. The key challenge of RNN-based methods is to model the spatial correlations of human joints. Du \etal \cite{du2015hierarchical} proposed a hierarchical RNN to process different parts of human body and fuse features in a hierarchical way. Liu \etal \cite{liu2016spatio} extended LSTM to spatio-temporal domain and proposed a tree-structure based traversal method to extract spatial features. Song \etal \cite{song2016end} leveraged attention mechanism to selectively focus on discriminative joints and key frames. 

%CNN-based methods typically organize skeleton data into a pseudo-image and utilize CNNs to extract features. Ke \etal \cite{ke2017new} proposed a new representation for 3D skeleton which provides useful spatial structural information. Li \etal \cite{li2017skeleton} proposed a skeleton transformer module to rearrange and select important skeleton joints adaptively. However, both RNN-based and CNN-based methods overlook the inherent graph structure of skeletons which restricts model capability of modeling correlations between human joints.

GCNs have been successfully adopted to skeleton-based action recognition \cite{liu2020disentangling,shi2019two,yan2018spatial,ye2020dynamic,zhao2019bayesian,tang2018deep} and most of them follow the feature update rule of \cite{kipf2016semi}. Due to the importance of topology (namely vertex connection relationship) in GCN, many GCN-based methods focus on topology modeling. According to the difference of topology, GCN-based methods can be categorized as follows: (1) According to whether the topology is dynamically adjusted during inference, GCN-based methods can be classified into static methods and dynamic methods. (2) According to whether the topology is shared across different channels, GCN-based methods can be classified into topology-shared methods and topology-non-shared methods.
\begin{figure*}[t]
	\centering
	\includegraphics[width=\linewidth]{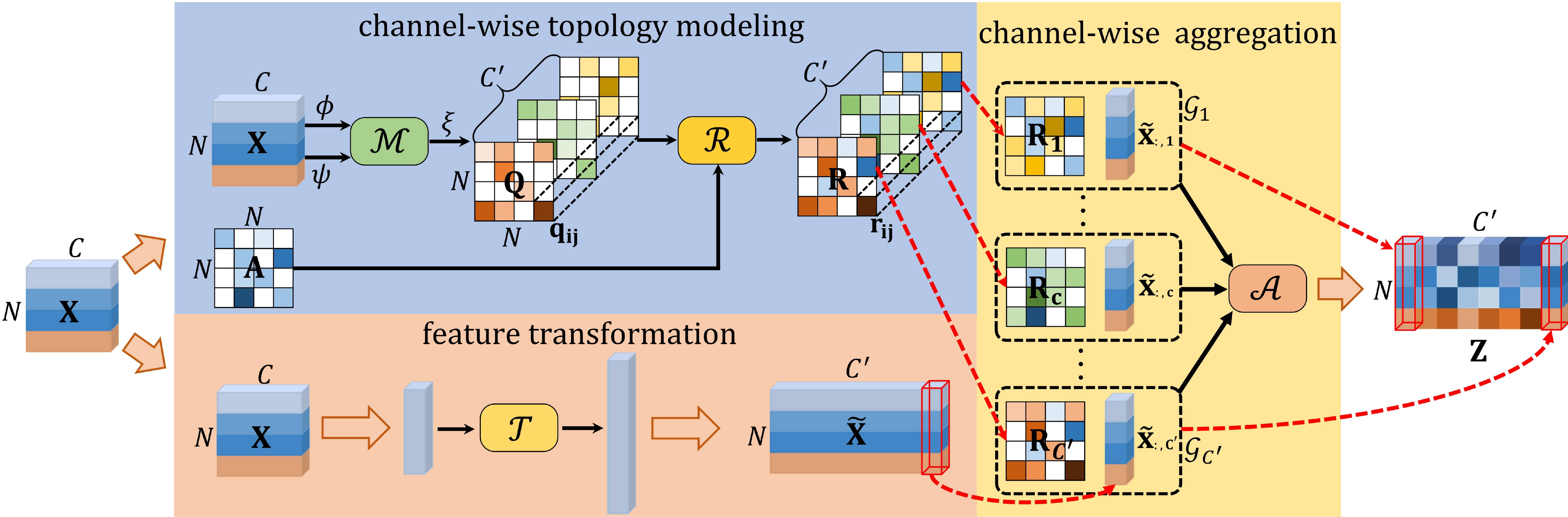} % Reduce the figure size so that it is slightly narrower than the column. Don't use precise values for figure width.This setup will avoid overfull boxes.
	\caption{Framework of the proposed channel-wise topology refinement graph convolution. The channel-wise topology modeling refines the trainable shared topology with inferred channel-specific correlations. The feature transformation aims at transforming input features into high-level representations. Eventually, the output feature is obtained by channel-wise aggregation.}
	\label{fig:framework}
	\vspace{-0.4cm}
\end{figure*}

\noindent \textbf{Static / Dynamic Methods.} For \textbf{static methods}, the topologies of GCNs keep fixed during inference. Yan \etal \cite{yan2018spatial} proposed an ST-GCN which predefines topology according to human body structure and the topology is fixed in both training and testing phase. Liu \etal \cite{liu2020disentangling} and Huang \etal \cite{huang2020spatio} introduced multi-scale graph topologies to GCNs to enable multi-range joint relationship modeling. For \textbf{dynamic methods}, the topologies of GCNs are dynamically inferred during inference. Li \etal \cite{li2019actional} proposed an A-links inference module to capture action-specific correlations. Shi \etal \cite{shi2019two} and Zhang \etal \cite{zhang2020semantics} enhanced topology learning with self-attention mechanism, which models correlation between two joints given corresponding features. These methods infer correlations between two joints with local features. Ye \etal \cite{ye2020dynamic} proposed a Dynamic GCN, where contextual features of all joints are incorporated to learn correlations between any pairs of joints. Compared with static methods, dynamic methods have stronger generalization ability due to dynamic topologies.

\noindent \textbf{Topology-shared / Topology-non-shared Methods.} For \textbf{topology-shared methods}, the static or dynamic topologies are shared in all channels. These methods force GCNs to aggregate features in different channels with the same topology, limiting the upper bound of model performance. Most GCN-based methods follow topology-shared manner, including aforementioned static methods \cite{huang2020spatio,liu2020disentangling,yan2018spatial} and dynamic methods \cite{li2019actional,shi2019two,ye2020dynamic,zhang2020semantics}. \textbf{Topology-non-shared methods} use different topologies in different channels or channel groups, which naturally overcome limitations of topology-shared methods. Cheng \etal \cite{cheng2020eccv} proposed a DC-GCN which sets individual parameterized topologies for different channel groups. However, the DC-GCN faces difficulty of optimization caused by excessive parameters when setting channel-wise topologies. To our best knowledge, topology-non-shared graph convolutions are rarely explored in the skeleton-based action recognition, and this work is the first to model dynamic channel-wise topologies. Note that our method also belongs to dynamic methods because topologies are dynamically inferred during inference.

%GCN-based methods utilize graph convolutions to learn correlations between human joints and achieve remarkable performance in skeleton-based action recognition. Yan \etal \cite{yan2018spatial} proposed an ST-GCN which for the first time utilizes spatial graph convolutions along with temporal convolutions to model motion patterns. Shi \etal \cite{shi2019two} proposed to learn the topology adaptively with self-attention mechanism and a parameterized adjacency matrix. They also use a two-stream ensemble with skeleton bone features to boost performance. Liu \etal \cite{liu2020disentangling} proposed a 3D graph convolution and a multi-scale aggregation scheme to enable direct information flow across the spatial-temporal graph and long-range correlation modeling respectively. These methods use the same topology for different channels, which limits the capability of correlation modeling. Cheng \etal \cite{cheng2020eccv} proposed a decoupling GCN which set individual parameterized topologies for different channel groups. However, modeling channel-wise topologies with decoupling GCN leads to a heavy model, which increases the difficulty of optimization and limits the model performance. Moreover, topologies in decoupling GCN are not adaptive to input samples, restricting the flexibility of modeling sample-dependent correlations. 

\section{Method}
In this section, we first define related notations and formulate conventional graph convolution. Then we elaborate our Channel-wise Topology Refinement Graph Convolution (CTR-GC) and mathematically analyze the representation capability of CTR-GC and other graph convolutions. Finally, we introduce the structure of our CTR-GCN.

\subsection{Preliminaries}
\noindent \textbf{Notations.} A human skeleton is represented as a graph with joints as vertices and bones as edges. The graph is denoted as $\mathcal{G=(V,E,X)}$, where $\mathcal{V}=\{v_1,v_2,......,v_N\}$ is the set of $N$ vertices. $\mathcal{E}$ is the edge set, which is formulated as an adjacency matrix $\mathbf{A} \in \mathbb{R}^{N \times N}$ and its element $a_{ij}$ reflects the correlation strength between $v_i$ and $v_j$. The neighborhood of $v_i$ is represented as $\mathcal{N}(v_i)=\{v_j|a_{ij}\neq 0\}$. $\mathcal{X}$ is the feature set of $N$ vertices, which is represented as a matrix $\mathbf{X}\in \mathbb{R}^{N \times C}$ and $v_i$'s feature is represented as $\mathbf{x_i} \in \mathbb{R}^C$. %Our CTR-GC aims at learning refined channel-wise topologies for each sample and updating the input $\mathbf{X}$ to a more correlation-aware representation $\mathbf{Z}$.% An action sample is represented as a sequence of human skeleton graphs of length $T$.
%we will first elaborate on our proposed Channel-wise Topology Refinement  Graph Convolution (CTR-GC), and then reveal the source of CTR-GC's superiority compared to other graph convolutions from the perspective of equivalent weights.

%For a 2D tensor $\mathbf{X}$, we represent (1) the vector lies in the i-th row as $\mathbf{x_{i}}$, (2) the vector lies in the j-th column as $\mathbf{x_{:,j}}$, (3) the element locates at the i-th row, j-th column as $x_{ij}$. For a 3D tensor $\mathbf{R} \in \mathbb{R}^{N\times N \times C}$ where the last dimension corresponds to channel dimension, we represent (1) the matrix lies in the c-th channel as $\mathbf{R^c}$, (2) the vector locates at the i-th row, j-th column as $\mathbf{r_{ij}}$, (3) the element locates at the the i-th row, j-th column, c-th channel as $r^c_{ij}$. 

\noindent \textbf{Topology-shared Graph Convolution.} The normal topology-shared graph convolution utilizes the weight $\mathbf{W}$ for feature transformation and aggregate representations of $v_i$'s neighbor vertices through $a_{ij}$ to update its representation $\mathbf{z_i}$, which is formulated as
\vspace{-0.2cm}
\begin{equation}
\vspace{-0.2cm}
\mathbf{z_i}=\sum_{v_j \in \mathcal{N}(v_i)}a_{ij}\mathbf{x_j}\mathbf{W} \label{eq:fa}
\end{equation}

For static methods, $a_{ij}$ is defined manually or set as trainable parameter. For dynamic methods, $a_{ij}$ is usually generated by the model depending on the input sample. %The above computation process can be decomposed into two steps, \ie, feature transformation by $\mathbf{W}$ and feature aggregation by $\mathbf{A}$.

\subsection{Channel-wise Topology Refinement Graph Convolution}

%The graph convolution described above uses the same topology (\ie, $\mathbf{A}$) for all channels, which restricts the capability of modeling correlations between vertices. To overcome this limitation, we propose a channel-wise topology refinement graph convolution.

The general framework of our CTR-GC is shown in Figure \ref{fig:framework}. We first transform input features into high-level features, then dynamically infer channel-wise topologies to capture pairwise correlations between input sample's joints under different types of motion features, and aggregate features in each channel with corresponding topology to get the final output. Specifically, our CTR-GC contains three parts: (1) Feature transformation which is done by transformation function $\mathcal{T}(\cdot)$; (2) Channel-wise topology modeling which consists of correlation modeling function $\mathcal{M}(\cdot)$ and refinement function $\mathcal{R}(\cdot)$; (3) Channel-wise aggregation which is completed by aggregation function $\mathcal{A}(\cdot)$. Given the input feature $\mathbf{X} \in \mathbb{R}^{N \times C}$, the output $\mathbf{Z} \in \mathbb{R}^{N \times C'}$ of CTR-GC is formulated as
\vspace{-0.2cm}
\begin{equation}
\vspace{-0.1cm}
\mathbf{Z}=\mathcal{A}\big( \mathcal{T}(\mathbf{X}), \mathcal{R}(\mathcal{M}(\mathbf{X}), \mathbf{A})\big),
\end{equation}
where $\mathbf{A} \in \mathbb{R}^{N \times N}$ is the learnable shared topology. Next, we introduce these three parts in detailed.

%Specifically, we use the correlation modeling function $\mathcal{M}$ and refinement function $\mathcal{R}$ to infer channel-wise topology, and then use

%Specifically, we use the transformation function $\mathcal{T}(\cdot)$ to transform input features into high-level features, and then infer channel-wise topologies through the channel-wise topology refinement process $\mathcal{A}(\cdot)$. Finally, we conduct channel-wise aggregation on high-level features with channel-wise topologies to obtain the output $\mathbf{Z}$.

%The input $\mathbf{X}$ is firstly transformed $\mathcal{T}(\cdot)$ is feature transformation function shared over all vertices, which transforms input feature to high-level feature. $\mathcal{R}(\cdot)$ corresponds to channel-wise topology refinement process which refines channel-specific topologies and $\mathcal{M}(\cdot)$ is correlation modeling function which infers channel-specific correlations between pairs of vertices. $\mathcal{A}(\cdot)$ is feature aggregation function which aggregates features into the finally representation $\mathbf{Z}$. Next, we introduce these three parts in detailed. %$\mathcal{A}(\cdot)$ should be symmetric function (\eg, summation) to achieve permutation invariance of vertices.

\noindent \textbf{Feature Transformation.} As shown in the orange block in Figure \ref{fig:framework}, feature transformation aims at transforming input features into high-level representations via $\mathcal{T}(\cdot)$. We adopt a simple linear transformation here as the topology-shared graph convolution, which is formulated as
\vspace{-0.2cm}
\begin{equation}
\vspace{-0.3cm}
\mathbf{\widetilde{X}}=\mathcal{T}(\mathbf{X})=\mathbf{XW},
\end{equation} 
where $\mathbf{\widetilde{X}} \in \mathbb{R}^{N \times C'}$ is the transformed feature and $\mathbf{W} \in \mathbb{R}^{C \times C'}$ is the weight matrix. Note that other transformations can also be used, \eg, multi-layer perceptron.

\noindent \textbf{Channel-wise Topology Modeling.} The channel-wise topology modeling is shown in the blue block in Figure \ref{fig:framework}. The adjacency matrix is used as shared topology for all channels and is learned through backpropagation. Moreover, we learn channel-specific correlations $\mathbf{Q} \in \mathbb{R}^{N \times N \times C'}$ to model specific relationships between vertices in $C'$ channels. Then the channel-wise topologies $\mathbf{R} \in \mathbb{R}^{N \times N \times C'}$ are obtained by refining the shared topology $\mathbf{A}$ with $\mathbf{Q}$.

Specifically, we first employ correlation modeling function $\mathcal{M}(\cdot)$ to model channel-wise correlations between vertices. To reduce computation cost, we utilize linear transformations $\phi$ and $\psi$ to reduce feature dimension before sending input features into $\mathcal{M}(\cdot)$. Given a pair of vertices $(v_i, v_j)$ and their corresponding features $(\mathbf{x_i}, \mathbf{x_j})$, we design two simple yet effective correlation modeling functions. The first correlation modeling function $\mathcal{M}_1(\cdot)$ is formulated as
\vspace{-0.16cm}
\begin{equation}
\vspace{-0.16cm}
\label{equ:cm1}
\mathcal{M}_1(\psi(\mathbf{x_i}), \phi(\mathbf{x_j}))=\sigma(\psi(\mathbf{x_i}) - \phi(\mathbf{x_j})),
\end{equation}
where $\sigma(\cdot)$ is activation function. $\mathcal{M}_1(\cdot)$ essentially calculates distances between $\psi(\mathbf{x_i})$ and $\phi(\mathbf{x_j})$ along channel dimension and utilizes the nonlinear transformations of these distances as channel-specific topological relationship between $v_i$ and $v_j$. The second correlation modeling function $\mathcal{M}_2(\cdot)$ is formulated as
\vspace{-0.16cm}
\begin{equation}
\vspace{-0.16cm}
\mathcal{M}_2(\psi(\mathbf{x_i}), \phi(\mathbf{x_j}))=MLP(\psi(\mathbf{x_i}) || \phi(\mathbf{x_j})),
\end{equation}
where $||$ is concatenate operation and MLP is multi-layer perceptron. We utilize MLP here due to its powerful fitting capability.

Based on the correlation modeling function, the channel-specific correlations $\mathbf{Q} \in \mathbb{R}^{N \times N \times C'}$ are obtained by employing linear transformation $\xi$ to raise the channel dimension, which is formulated as
\vspace{-0.16cm}
\begin{equation}
\vspace{-0.16cm}
\mathbf{q_{ij}}=\xi\Big(\mathcal{M}\big(\psi(\mathbf{x_i}), \phi(\mathbf{x_j})\big)\Big),\ i,j \in \{1,2,\cdots,N\},
\label{equ:qij}
\end{equation}
where $\mathbf{q_{ij}} \in \mathbb{R}^{C'}$ is a vector in $\mathbf{Q}$ and reflects the channel-specific topological relationship between $v_i$ and $v_j$. Note that $\mathbf{Q}$ is not forced to be symmetric, \ie, $\mathbf{q_{ij}} \neq\mathbf{q_{ji}}$, which increases the flexibility of correlation modeling.

Eventually, the channel-wise topologies $\mathbf{R} \in \mathbb{R}^{N \times N \times C'}$ are obtained by refining the shared topology $\mathbf{A}$ with channel-specific correlations $\mathbf{Q}$:
\vspace{-0.16cm}
\begin{equation}
\vspace{-0.16cm}
\mathbf{R}= \mathcal{R}(\mathbf{Q}, \mathbf{A}) =\mathbf{A} + \alpha \cdot \mathbf{Q},
\end{equation}
where $\alpha$ is a trainable scalar to adjust the intensity of refinement. The addition is conducted in a broadcast way where $\mathbf{A}$ is added to each channel of $\alpha \times \mathbf{Q}$.

\noindent \textbf{Channel-wise Aggregation.} Given the refined channel-wise topologies $\mathbf{R}$ and high-level features $\mathbf{\widetilde{X}}$, CTR-GC aggregates features in a channel-wise manner. Specifically, CTR-GC constructs a channel-graph for each channel with corresponding refined topology $\mathbf{R_c} \in \mathbb{R}^{N \times N}$ and feature $\mathbf{\tilde{x}_{:,c}} \in \mathbb{R}^{N \times 1}$, where $\mathbf{R_c}$ and $\mathbf{\tilde{x}_{:,c}}$ are respectively from $\mathbf{c}$-th channel of $\mathbf{R_c}$ and $\mathbf{\widetilde{X}}$ ($c\in\{ 1,\cdots,C'\}$). Each channel-graph reflects relationships of vertices under a certain type of motion feature. Consequently, feature aggregation is performed on each channel-graph, and the final output $\mathbf{Z}$ is obtained by concatenating the output features of all channel-graphs, which is formulated as
\vspace{-0.15cm}
\begin{equation}
\vspace{-0.15cm}
\label{equ:hammada}
\mathbf{Z}=\mathcal{A}(\mathbf{\widetilde{X},R})=[\mathbf{R_1} \mathbf{\tilde{x}_{:,1}} ||\mathbf{R_2} \mathbf{\tilde{x}_{:,2}}||\cdots||\mathbf{R_{C'}} \mathbf{\tilde{x}_{:,C'}}],
\end{equation}
where $||$ is concatenate operation. During the whole process, the inference of channel-specific correlations $\mathbf{Q}$ relies on input samples as shown in Equation \ref{equ:qij}. Therefore, the proposed CTR-GC is a dynamic graph convolution and it adaptively varies with different input samples.

\subsection{Analysis of Graph Convolutions}
We analyze the representation capability of different graph convolutions by reformulating them into a unified form and comparing them with dynamic convolution \cite{chen2020dynamic,yang2019condconv} employed in CNNs.

We first recall dynamic convolution which enhances vanilla convolution with dynamic weights. In dynamic convolution, each neighbor pixel $p_j$ of the center pixel $p_i$ has a corresponding weight in the convolution kernel, and the weight can be dynamically adjusted according to different input samples, which makes the dynamic convolution have strong representation ability. The dynamic convolution can be formulated as
\vspace{-0.15cm}
\begin{equation}
\vspace{-0.15cm}
\mathbf{z_i^k}=\sum_{p_j \in \mathcal{N}(p_i)}\mathbf{x_j^k}\mathbf{W_j^k}, \label{equ:conv}
\end{equation}
where $\mathbf{k}$ indicates the index of input sample. $\mathbf{x_j^k}$ and $\mathbf{z_i^k}$ are the input feature of $p_j$ and the output feature of $p_i$ of the $\mathbf{k}$-th sample. $\mathbf{W_j^k}$ is the dynamic weight. 

%In dynamic convolution, each $p_j$ corresponds to an individual weight $\mathbf{W_j^k}$ and the weight can be dynamically adjusted according to input sample, which makes dynamic convolution has strong representation capability. 

Due to the irregular structure of the graph, the correspondence between neighbor vertices and weights is difficult to establish. Thus, graph convolutions (GCs) degrade convolution weights into adjacency weights (\ie, topology) and weights shared in the neighborhood. However, sharing weights in the neighborhood limits representation capability of GCs. To analyze the gap of representation ability between different GCs and dynamic convolution, we integrate adjacency weights and weights shared in the neighborhood into a generalized weight matrix $\mathbf{E^k_{ij}}$. Namely, we formulate all GCs in the form of $\mathbf{z_i^k}=\sum_{v_j \in \mathcal{N}(v_i)}\mathbf{x_j^k}\mathbf{E^k_{ij}}$ where $\mathbf{E^k_{ij}}$ is generalized weight. We classified GCs into four categories as mentioned before.%we reformulate them and derive their equivalent convolution weights as follows.

\noindent \textbf{Static Topology-shared GCs.} In static topology-shared GCs, the topologies keep fixed for different samples and are shared across all channels, which can be formulated as 
\begin{equation}
\mathbf{z_i^k}=\sum_{v_j \in \mathcal{N}(v_i)}a_{ij}\mathbf{x_j^k}\mathbf{W}=\sum_{v_j \in \mathcal{N}(v_i)}\mathbf{x_j^k}(a_{ij}\mathbf{W}), \label{equ:gc_equivalent}
\end{equation}
where  $a_{ij}\mathbf{W}$ is the generalized weight of static topology-shared GC. From Equation \ref{equ:conv} and \ref{equ:gc_equivalent}, it can be seen that the difference between dynamic convolution and static topology-shared GC lies in their (generalized) weights. Specifically, the weights of dynamic convolution $\mathbf{W_j^k}$ is individual for each $j$ and $k$, while generalized weights of static topology-shared GC is subject to following constraints:

\noindent \textit{Constraint 1: $\mathbf{E^{k_1}_{ij}}$ and $\mathbf{E^{k_2}_{ij}}$ are forced to be same.}

\noindent \textit{Constraint 2: $\mathbf{E^k_{ij_1}}$ and $\mathbf{E^k_{ij_2}}$ differ by a scaling factor.}

Note that $\mathbf{k_1}, \mathbf{k_2}$ are different sample indices and $\mathbf{j_1},\mathbf{j_2}$ are different neighbor vertex indices. These constraints cause the gap of representation ability between static topology-shared GCs and dynamic convolutions. Note that we concentrate on the neighborhood rooted at $v_i$ and do not consider the change of $v_i$ for simplicity.

%Note that for topology-non-shared graph convolution, we just analyze ones containing channel-wise topologies, because graph convolution with channel-wise topology can degenerate into other graph convolutions with non-shared topology (\ie, group-wise topology).

\noindent \textbf{Dynamic topology-shared GCs.} Compared with static topology-shared GCs, the dynamic ones infer topologies dynamically and thus have better generalization ability. The formulation of dynamic topology-shared GCs is
\vspace{-0.15cm}
\begin{equation}
\vspace{-0.15cm}
\mathbf{z_i^k}=\sum_{v_j \in \mathcal{N}(v_i)}a_{ij}^k\mathbf{x_j^k}\mathbf{W}\\
=\sum_{v_j \in \mathcal{N}(v_i)}\mathbf{x_j^k}(a_{ij}^k\mathbf{W}),
\label{equ:dynamic_shared_gc}
\end{equation}
where $a_{ij}^k$ is dynamic topological relationship between $v_i$, $v_j$ and depends on input sample. It can be seen that the generalized weights of dynamic topology-shared GCs still suffer from \textit{Constraint 2} but relax \textit{Constraint 1} into the following constraint:

\noindent \textit{Constraint 3: $\mathbf{E^{k_1}_{ij}}$, $\mathbf{E^{k_2}_{ij}}$ differ by a scaling factor.}

\noindent \textbf{Static topology-non-shared GCs.} This kind of GCs utilize different topologies for different channels (groups). Here we just analyze static GCs with channel-wise topologies because it is the most generalized form of static topology-non-shared GCs and can degenerate into others, \eg, static group-wise-topology GCs. The specific formulation is
\vspace{-0.15cm}
\begin{align}
\mathbf{z_i^k}&=\sum_{v_j \in \mathcal{N}(v_i)}\mathbf{p_{ij}}\odot(\mathbf{x_j^k}\mathbf{W}) \label{equ:pagc_o}\\
&=\sum_{v_j \in \mathcal{N}(v_i)}\mathbf{x_j^k}\big([p_{ij1}\mathbf{w_{:,1}},\cdots, p_{ijC'}\mathbf{w_{:,C'}}]\big) \label{equ:pagc},
\vspace{-0.5cm}
\end{align}
where $\odot$ is element-wise multiplication and $\mathbf{p_{ij}}\in \mathbb{R}^{C'}$ is channel-wise topological relationship between $v_i$, $v_j$. $p_{ijc}$ is the $c$-th element of $\mathbf{p_{ij}}$. $\mathbf{w_{:,c}}$ is the $c$-th column of $\mathbf{W}$. (We omit the derivation of Equation \ref{equ:pagc_o} and \ref{equ:pagc} for clarity. The details can be found in the supplementary materials.) From Equation \ref{equ:pagc}, we observe that generalized weights of this kind of GCs suffer from \textit{Constraint 1} due to static topology but relax \textit{Constraint 2} into the following constraint:

\noindent \textit{Constraint 4: Different corresponding columns of $\mathbf{E^k_{ij_1}}$ and $\mathbf{E^k_{ij_2}}$ differ by different scaling factors.}

\noindent \textbf{Dynamic topology-non-shared GCs.} The only difference between static topology-non-shared GCs and dynamic topology-non-shared GCs is that dynamic topology-non-shared GCs infers non-shared topologies dynamically, thus dynamic topology-non-shared GCs can be formulated as
\vspace{-0.15cm}
\begin{equation}
\vspace{-0.15cm}
\mathbf{z_i^k}=\sum_{v_j \in \mathcal{N}(v_i)}\mathbf{x_j^k}\big([r_{ij1}^k\mathbf{w_{:,1}},\cdots, r_{ijC'}^k\mathbf{w_{:,C'}}]\big), \label{equ:dynamic_non_shared}
\end{equation}
where $r_{ijc}^k$ is the $\mathbf{k}$-th sample's dynamic topological relationship between $v_i$, $v_j$ in the $c$-th channel. Obviously, generalized weights of dynamic topology-non-shared graph convolution relax both \textit{Constraint 1} and \textit{2}. Specifically, it relaxes \textit{Constraint 2} into \textit{Constraint 4} and relaxes \textit{Constraint 1} into the following constraint:

\noindent \textit{Constraint 5: Different corresponding columns of $\mathbf{E^{k_1}_{ij}}$ and $\mathbf{E^{k_2}_{ij}}$ differ by different scaling factors.}

\begin{table}
	\begin{center}
		\resizebox{.48\textwidth}{!}{
		\begin{tabular}{c c |c c c c c |c}
			\hline
			\multicolumn{2}{c|}{\textbf{Topology}} & \multicolumn{5}{c|}{\textbf{Constraints}} & \multirow{2}{*}{\textbf{Instance}}\\

			 \multicolumn{1}{c}{Non-shared} & Dynamic & {\color{red}\textbf{1}} & {\color{green}\textbf{2}}& {\color{green}\textbf{3}}& {\color{blue}\textbf{4}}& {\color{blue}\textbf{5}} &\\
			 \hline \hline
			\xmark & \xmark & \cmark &\cmark & & & & ST-GC\cite{yan2018spatial}\\
			\xmark & \cmark &  &\cmark &\cmark & & & AGC \cite{shi2019two}, Dy-GC\cite{ye2020dynamic}\\
			\cmark & \xmark & \cmark & & &\cmark& &DC-GC \cite{cheng2020eccv}\\
			\cmark & \cmark &  & & &\cmark&\cmark& \textbf{CTR-GC (ours)}\\
			\hline
		\end{tabular}}
	\end{center}
	\vspace{-0.2cm}
	\caption{Constraints on different categories of graph convolutions and corresponding instances. The number 1-5 correspond to five constraints. {\color{red}Red}, {\color{green}Green} and {\color{blue}Blue} respectively indicate the relatively {\color{red}High}, {\color{green}Mid} and {\color{blue}Low} constraint strength.}
	\label{tab:constraints}
	\vspace{-0.3cm}
\end{table}

We conclude different categories of graph convolutions and their constraints in Table \ref{tab:constraints}. It can be seen that dynamic topology-non-shared GC is the least constrained. Our CTR-GC belongs to dynamic topology-non-shared GC and Equation \ref{equ:hammada} can be reformulated to Equation \ref{equ:dynamic_non_shared}, indicating that theoretically CTR-GC has stronger representation capability than previous graph convolutions \cite{cheng2020eccv,shi2019two,yan2018spatial, ye2020dynamic}. The specific reformulation is shown in supplemental materials.

\subsection{Model Architecture}
\label{sec:model_architecture}
\begin{figure}[t]
	\centering
	\includegraphics[width=\columnwidth]{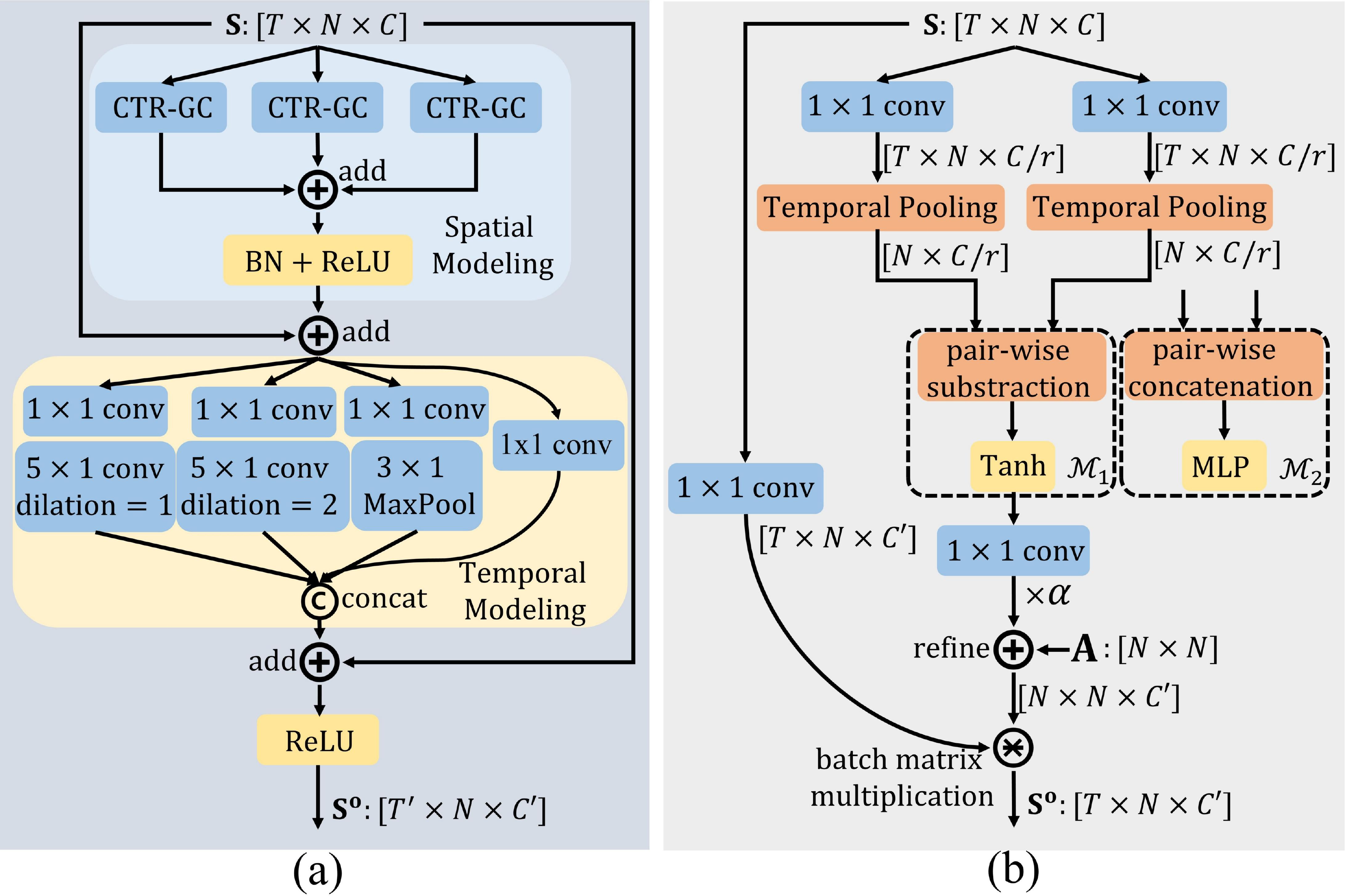} % Reduce the figure size so that it is slightly narrower than the column. Don't use precise values for figure width.This setup will avoid overfull boxes.
	\caption{(a) The basic block of our CTR-GCN. (b)CTR-GC with correlation modeling function $\mathcal{M}_1(\cdot)$ or $\mathcal{M}_2(\cdot)$.}
	\label{fig:gc_instance}
	\vspace{-0.5cm}
\end{figure}

Based on CTR-GC, we construct a powerful graph convolutional network CTR-GCN for skeleton-based action recognition. We set the neighborhood of each joint as the entire human skeleton graph, which is proved to be more effective in this task by previous work \cite{cheng2020skeleton,shi2019two}. The entire network consists of ten basic blocks, followed by a global average pooling and a softmax classifier to predict action labels. The number of channels for ten blocks are 64-64-64-64-128-128-128-256-256-256. Temporal dimension is halved at the 5-th and 8-th blocks by strided temporal convolution. The basic block of our CTR-GCN is shown in Figure \ref{fig:gc_instance} (a). Each block mainly consists of a spatial modeling module, a temporal modeling module and residual connections. 

\noindent \textbf{Spatial Modeling.} In a spatial modeling module, we use three CTR-GCs in parallel to extract correlations between human joints and sum up their results as output. For clarity, an instance of CTR-GC with $\mathcal{M}_1(\cdot)$ is illustrated in Figure \ref{fig:gc_instance} (b). Our CTR-GC is designed to extract features of a graph with input feature $\mathbf{X} \in \mathbb{R}^{N \times C}$. To adopt CTR-GC to a skeleton graph sequence $\mathbf{S} \in \mathbb{R}^{T\times N\times C}$, we pool $\mathbf{S}$ along temporal dimension and use pooled features to infer channel-wise topologies. Specifically, CTR-GC first utilizes $\phi$ and $\psi$ with reduction rate $r$ to extract compact representations. Then temporal pooling is used to aggregate temporal features. After that, CTR-GC conducts pair-wise subtraction and activation following Equation \ref{equ:cm1}. The channel dimension of activation is then raised with $\xi$ to obtain channel-specific correlations, which are used to refine the shared topology $\mathbf{A}$ to obtain channel-wise topologies. Eventually, channel-wise aggregation (implemented by batch matrix multiplication) is conducted in each skeleton graph to obtain the output representation $\mathbf{S^o}$.

\noindent \textbf{Temporal Modeling.} To model actions with different duration, we design a multi-scale temporal modeling module following \cite{liu2020disentangling}. The main difference is that we use fewer branches for that too many branches slow down inference speed. As shown in Figure \ref{fig:gc_instance} (a), this module contains four branches, each containing a $1 \times 1$ convolution to reduce channel dimension. The first three branches contain two temporal convolutions with different dilations and one MaxPool respectively following $1\times 1$ convolution. The results of four branches are concatenated to obtain the output.

%\noindent \textbf{Specialization.} As for skeleton-based action recognition, the input $\mathbf{X} \in \mathbb{R}^{C\times T\times V}$ is a sequence of human skeleton which has $T$ frames and $V$ joints per frame. The proposed PAGC are utilized to extract spatial information of human skeleton sequences and the receptive field is set to the whole skeleton graph to extract global information. An instance of PAGC with correlation modeling function $\mathcal{M}_1(\cdot)$ is illustrated in Figure \ref{fig:gc_instance} (a). Given the input $\mathbf{X}$, we first utilize two $1\times 1$ convolutions with reduction rate $r$ to extract compact presentations. Then temporal pooling are used to aggregate temporal features across different time steps. After that, we transpose one representation and subtract it with another in a broadcasting way which follows the Equation \ref{equ:cm1}. This subtracting result is activated by Tanh function and the channel is raised to output channel with a $1\times 1$ convolution to obtain pattern-wise correlations. Eventually, we conduct pattern-wise aggregation which is implemented with batch matrix multiplication to obtain the output representation.

\section{Experiments}

\subsection{Datasets}
\noindent \textbf{NTU RGB+D.} NTU RGB+D \cite{shahroudy2016ntu} is a large-scale human action recognition dataset containing 56,880 skeleton action sequences. The action samples are performed by 40 volunteers and categorized into 60 classes. Each sample contains an action and is guaranteed to have at most 2 subjects, which is captured by three Microsoft Kinect v2 cameras from different views concurrently. The authors of this dataset recommend two benchmarks: (1) cross-subject (X-sub): training data comes from 20 subjects, and testing data comes from the other 20 subjects. (2) cross-view (X-view): training data comes from camera views 2 and 3, and testing data comes from camera view 1.

\noindent \textbf{NTU RGB+D 120.} NTU RGB+D 120 \cite{liu2019ntu} is currently the largest dataset with 3D joints annotations for human action recognition, which extends NTU RGB+D with additional 57,367 skeleton sequences over 60 extra action classes. Totally 113,945 samples over 120 classes are performed by 106 volunteers, captured with three cameras views. This dataset contains 32 setups, each denoting a specific location and background. The authors of this dataset recommend two benchmarks: (1) cross-subject (X-sub): training data comes from 53 subjects, and testing data comes from the other 53 subjects. (2) cross-setup (X-setup): training data comes from samples with even setup IDs, and testing data comes from samples with odd setup IDs.

\noindent \textbf{Northwestern-UCLA.} Northwestern-UCLA dataset \cite{wang2014cross} is captured by three Kinect cameras simultaneously from multiple viewpoints. It contains 1494 video clips covering 10 action categories. Each action is performed by 10 different subjects. We follow the same evaluation protocol in \cite{wang2014cross}: training data from the first two cameras, and testing data from the other camera.
\subsection{Implementation Details}
All experiments are conducted on one RTX 2080 TI GPU with the PyTorch deep learning framework. Our models are trained with SGD with momentum 0.9, weight decay 0.0004. The training epoch is set to 65 and a warmup strategy \cite{he2016deep} is used in the first 5 epochs to make the training procedure more stable. Learning rate is set to 0.1 and decays with a factor 0.1 at epoch 35 and 55. For NTU RGB+D and NTU RGB+D 120, the batch size is 64, each sample is resized to 64 frames, and we adopt the data pre-processing in \cite{zhang2020semantics}. For Northwestern-UCLA, the batch size is 16, and we adopt the data pre-processing in \cite{cheng2020skeleton}.

\subsection{Ablation Study}
In this section, we analyze the proposed channel-wise topology refinement graph convolution and its configuration on the X-sub benchmark of the NTU RGB+D 120 dataset.

\noindent \textbf{Effectiveness of CTR-GC.}
\begin{table}
	\begin{center}
		\begin{tabular}{l c l}
			\hline
			\textbf{Methods} & \textbf{Param.} & \ \ \textbf{Acc (\%)} \\
			\hline\hline
			Baseline & 1.22M & \ \ \  83.4 \\
			\hline
			+2 CTR-GC & 1.26M & \ \ \ 84.2 $^{\uparrow 0.8}$\\
			+5 CTR-GC & 1.35M & \ \ \ 84.7 $^{\uparrow 1.3}$\\
			\hline
			CTR-GCN w/o Q & 1.22M & \ \ \ 83.7 $^{\uparrow 0.3}$ \\
			CTR-GCN w/o A & 1.46M & \ \ \ 84.0 $^{\uparrow 0.6}$\\
			\hline
			CTR-GCN & 1.46M & \ \ \ \textbf{84.9 $^{\mathbf{\uparrow} \mathbf{1.5}}$} \\
			\hline
		\end{tabular}
	\end{center}
	\vspace{-0.2cm}
	\caption{Comparisons of accuracies when adding CTR-GCs gradually and removing A or Q from CTR-GCN.}
	\vspace{-0.3cm}
	\label{tab:pagc_acc}
\end{table}
We employ ST-GCN \cite{yan2018spatial} as the baseline, which belongs to static topology-shared graph convolution and the topology is untrainable. We further add residual connections in ST-GCN as our basic block and replace its temporal convolution with temporal modeling module described in Section \ref{sec:model_architecture} for fair comparison.

The experimental results are shown in Table \ref{tab:pagc_acc}. First, we gradually replace GCs with CTR-GCs (shown in Figure \ref{fig:gc_instance} (b) and $r=8$) in the baseline. We observe that accuracies increase steadily and the accuracy is substantially improved when all GCs are replaced by CTR-GCs (CTR-GCN), which validates the effectiveness of CTR-GC. 

Then we validate effects of the shared topology A and the channel-specific correlations Q respectively by removing either of them from CTR-GCN. CTR-GCN w/o Q shares a trainable topology across different channels. We observe that its performance drops 1.2\% compared with CTR-GCN, indicating the importance of modeling channel-wise topologies. The performance of CTR-GCN w/o A drops 0.9\%, confirming that it's hard to model individual topology for each channel directly and topology refinement provides an effective approach to solve this problem.

\noindent \textbf{Configuration Exploration.}
\begin{table}
	\begin{center}
		\begin{tabular}{c c c c c l}
			\hline
			\textbf{Methods} & $\boldsymbol{\mathcal{M}}$ & $\boldsymbol{r}$ & $\boldsymbol{\sigma}$ & \textbf{Param.} & \ \ \textbf{Acc (\%)}\\
			\hline\hline
			Baseline & - & - & -  & 1.21M & \ \ \ 83.4 \\
			\hline
			A & $\mathcal{M}_1^+$ & 8 & Tanh &  1.46M & \ \ \ \textbf{84.9}$^{\uparrow \mathbf{1.5}}$\\
			B & $\mathcal{M}_1$ & 8 & Tanh &  1.46M & \ \ \ \textbf{84.9}$^{\uparrow \mathbf{1.5}}$\\
			C & $\mathcal{M}_2$ & 8 & Tanh & 1.48M & \ \ \ 84.8$^{\uparrow 1.4}$\\
			\hline
			D & $\mathcal{M}_1$ &4 & Tanh &  1.69M & \ \ \ 84.8$^{\uparrow 1.4}$\\
			E & $\mathcal{M}_1$ &16 & Tanh & 1.34M & \ \ \ 84.5$^{\uparrow 1.1}$\\
			\hline
			F & $\mathcal{M}_1$ &8 & Sig & 1.46M & \ \ \ 84.6$^{\uparrow 1.2}$\\
			G & $\mathcal{M}_1$ &8 & ReLU & 1.46M & \ \ \ 84.8$^{\uparrow 1.4}$\\
			\hline
		\end{tabular}
	\end{center}
	\vspace{-0.2cm}
	\caption{Comparisons of the validation accuracy of CTR-GC with different settings.}
	\vspace{-0.5cm}
	\label{tab:config}
\end{table}
We explore different configurations of CTR-GC, including the choice of correlation modeling functions $\mathcal{M}$, the reduction rate $r$ of $\phi$ and $\psi$, activation function $\sigma$ of correlation modeling function. As shown in Table \ref{tab:config}, we observe that models under all configurations outperform the baseline, confirming the robustness of CTR-GC. (1) Comparing models A, B and C, we find models with different correlation modeling functions all achieve good performance, which indicates that channel-wise topology refinement is a generic idea and is compatible with many different correlation modeling functions ($\mathcal{M}_1^+$ replaces the subtraction in $\mathcal{M}_1$ with addition). (2) Comparing models B, D and E, we find models with $r=4,8$ (models B, D) achieve better results and the model with $r=8$ (model B) performs better slightly with fewer parameters. Model E with $r=16$ performs worse because too few channels are used in correlation modeling function, which is not sufficient to effectively model channel-specific correlations. (3) Comparing models B, F and G, Sigmoid and ReLU perform worse than Tanh and we argue that non-negative output values of Sigmoid and ReLU constrains the flexibility of correlation modeling. Considering performance and efficiency, we choose model B as our final model.

%To verify the effect of model size, we reduce the model size by removing two CTR-GCs in spatial modeling module in Figure \ref{fig:gc_instance} (a) while decreasing number of channels in convolutions (model F), and scale up the model size by increasing number of channels in convolutions (model G). Model F outperforms the baseline by 1.7\% with about 40\% parameters, which proves the powerful representation capability of CTR-GC. By scaling up the model size, model G achieves the best performance and exceeds the baseline by 3.1\%.
\subsection{Comparison with Other GCs}
\begin{table}
	\begin{center}
		\resizebox{.48\textwidth}{!}{
		\begin{tabular}{c c l c c c}
			\hline
			\multicolumn{2}{c}{\textbf{Topology}} & \multirow{2}{*}{\textbf{Methods}} & \multirow{2}{*}{\textbf{Param.}} & \multirow{2}{*}{\textbf{FLOPs}} & \multirow{2}{*}{\textbf{Acc (\%)}} \\
			Non-share & Dynamic & & & & \\
			\hline\hline
			\xmark & \xmark &ST-GC \cite{yan2018spatial}& 1.22M & \textasciitilde1.65G & 83.4 \\
			%\cline{3-6}
			\xmark &\cmark &AGC \cite{shi2019two} & 1.55M & \textasciitilde2.11G & 83.9\\
			%\cline{3-6}
			\xmark & \cmark &Dy-GC \cite{ye2020dynamic} & 1.73M & \textasciitilde1.66G & 83.9\\
			%\cline{3-6}
			\cmark & \xmark &DC-GC \cite{cheng2020eccv} & 1.51M & \textasciitilde1.65G &84.2 \\
			%\cline{3-6}
			\cmark& \xmark& DC-GC*\cite{cheng2020eccv} & 3.37M & \textasciitilde1.65G & 84.0 \\
			%\cline{3-6}
			\cmark&\cmark&CTR-GC & 1.46M & \textasciitilde1.97G &\textbf{84.9} \\
			\hline
		\end{tabular}}
	\end{center}
	\vspace{-0.2cm}
	\caption{Comparisons of CTR-GC with other graph convolutions. The first two columns show the categories of graph convolutions.}
	\vspace{-0.4cm}
	\label{tab:compare_other_gcns}

\end{table}
%\begin{figure}[t]
%	\centering
%	\includegraphics[width=\columnwidth]{figure/acc_v1.3.jpg} % Reduce the figure size so that it is slightly narrower than the column. Don't use precise values for figure width.This setup will avoid overfull boxes.
%	\caption{Comparison of classification accuracy of different graph convolutions on hard action classes.} 
%	\vspace{-0.3cm}
%	\label{fig:acc}
%\end{figure}
In order to validate the effectiveness of our CTR-GC, we compare performance, parameters and computation cost of CTR-GC against other graph convolutions in Table \ref{tab:compare_other_gcns}. Specifically, we keep the backbone of the baseline model and only replace graph convolutions for fair comparison. Note that DC-GC split channels into 16 groups and set a trainable adjacency matrix for each group, while DC-GC* set a trainable adjacency matrix for each channel. From Table \ref{tab:compare_other_gcns}, we observe that (1) On the whole, topology-non-shared methods achieve better performance than topology-shared methods, and dynamic methods perform better than static methods, indicating the importance of modeling non-shared topologies and dynamic topologies; (2) Compared with DC-GC, DC-GC* performs worse while has much more parameters, confirming that it's not effective to model channel-wise topologies with parameterized adjacency matrices alone; (3) CTR-GC outperforms DC-GC* by 0.9\%, proving that our refinement approach is effective to model channel-wise topologies. Moreover, our CTR-GC introduces little extra parameters and computation cost compared with other graph convolutions.

%We further analyze the performance of different graph convolutions on action classes with low accuracy, \ie, ``staple book", ``count money", ``play with phone" and ``cut nails", ``playing magic cube" and ``open bottle". These actions mainly involve subtle interactions between fingers, making them difficult to be recognized correctly. As shown in Figure \ref{fig:acc}, CTR-GC outperforms other graph convolutions on all classes. Especially, CTR-GC exceeds other methods at least by 7.03\% and 4.36\% on ``cut nails" and ``open bottle" respectively, showing that, compared with other GCs, our CTR-GC can effectively extract features of subtle interactions and classify them more accurately.

\begin{figure}[t]
	\centering
	\includegraphics[width=\columnwidth]{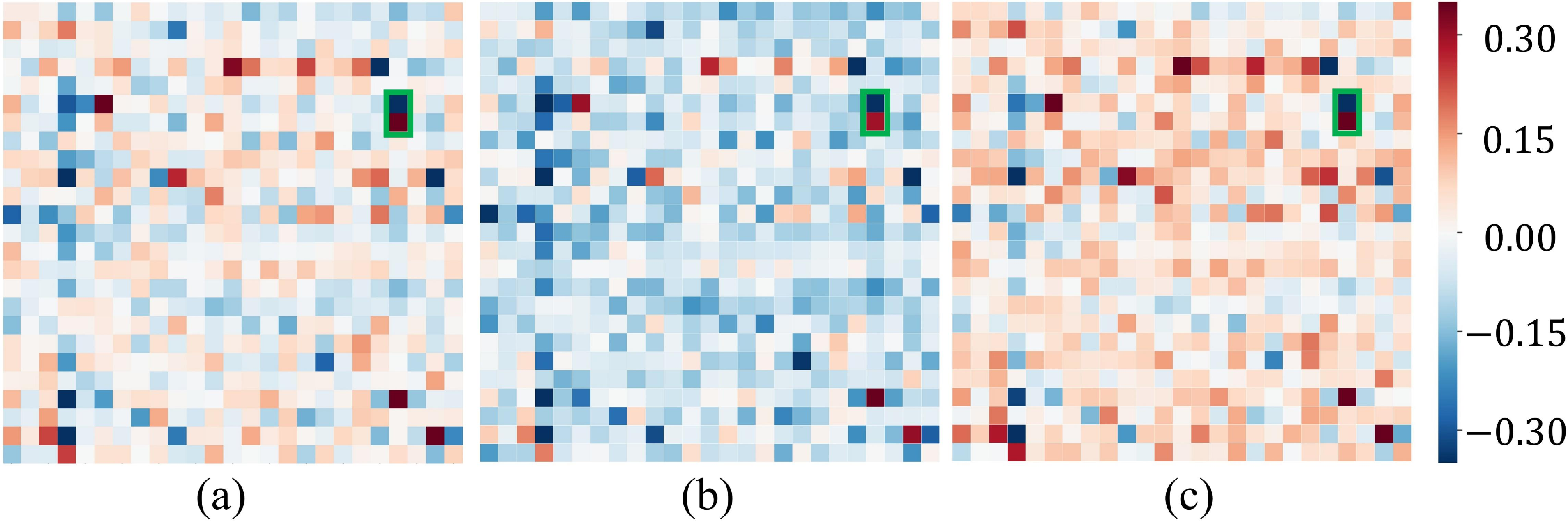} % Reduce the figure size so that it is slightly narrower than the column. Don't use precise values for figure width.This setup will avoid overfull boxes.
	\caption{(a) The shared topology. (b) and (c) The refined channel-wise topologies of different channels.}
	\vspace{-0.2cm}
	\label{fig:correlation}
\end{figure}

\subsection{Visualization of Learned Topologies}
We illustrate the shared topology and refined channel-wise topologies of an action sample ``typing on the keyboard" in Figure \ref{fig:correlation}. The values close to 0 indicate weak relationships between joints and vice versa. We observe that (1) the shared topology is different from refined channel-wise topologies, indicating that our method can effectively refine the shared topology. (2) the refined channel-wise topologies are different, demonstrating that our method can learn individual topologies depending on specific motion features for different channels. (3) Some correlations are consistently strong in all channels, indicating that these joint pairs are strongly relevant in general, \eg, the correlation between left elbow and left-hand tip (blue square in the green box), and the correlation between left-hand tip and left wrist (red square in the green box). It's reasonable for ``typing on the keyboard" where main motion happens on hands.

\subsection{Comparison with the State-of-the-Art}

\begin{table}
	\begin{center}
		\begin{tabular}{l c c}
			\hline
			\multirow{2}{*}{\textbf{Methods}} & \multicolumn{2}{c}{\textbf{NTU-RGB+D 120}}\\
			  & \textbf{X-Sub (\%)} & \textbf{X-Set (\%)} \\
			\hline\hline
			ST-LSTM\cite{liu2016spatio} & 55.7 & 57.9\\
			GCA-LSTM\cite{liu2017skeleton} & 61.2 & 63.3\\
			RotClips+MTCNN\cite{ke2018learning} & 62.2 & 61.8\\
			%Body Pose Evolution Map\cite{liu2018recognizing} & 64.6 & 66.9 & -\\
			\hline
			SGN\cite{zhang2020semantics} & 79.2 & 81.5\\
			2s-AGCN\cite{shi2019two} & 82.9 & 84.9\\
			Shift-GCN\cite{cheng2020skeleton} & 85.9 & 87.6\\
			DC-GCN+ADG\cite{cheng2020eccv} & 86.5 & 88.1\\
			MS-G3D\cite{liu2020disentangling} & 86.9 & 88.4\\
			PA-ResGCN-B19 \cite{song2020stronger} & 87.3 & 88.3 \\
			Dynamic GCN \cite{ye2020dynamic} & 87.3 & 88.6 \\
			\hline\hline
			CTR-GCN (Bone Only) & 85.7 & 87.5 \\
			CTR-GCN (Joint+Bone) & 88.7 & 90.1 \\
			\textbf{CTR-GCN} & \textbf{88.9} & \textbf{90.6}\\
			\hline
		\end{tabular}
	\end{center}
	\vspace{-0.3cm}
	\caption{Classification accuracy comparison against state-of-the-art methods on the NTU RGB+D 120 dataset.}
	\vspace{-0.5cm}
	\label{tab:ntu120}
\end{table}

\begin{table}
	\begin{center}
		\begin{tabular}{l c c}
			\hline
			\multirow{2}{*}{\textbf{Methods}} & \multicolumn{2}{c}{\textbf{NTU-RGB+D}}\\
			& \textbf{X-Sub (\%)} & \textbf{X-View (\%)} \\
			\hline\hline
			Ind-RNN\cite{li2018independently} & 81.8 & 88.0 \\
			HCN\cite{li2018co} & 86.5 & 91.1\\
			\hline
			ST-GCN\cite{yan2018spatial} & 81.5 & 88.3\\
			2s-AGCN\cite{shi2019two} & 88.5 & 95.1\\
			SGN\cite{zhang2020semantics} & 89.0 & 94.5\\
			AGC-LSTM\cite{si2019attention} & 89.2 & 95.0\\
			DGNN\cite{shi2019skeleton} & 89.9 & 96.1\\
			Shift-GCN\cite{cheng2020skeleton} & 90.7 & 96.5\\
			DC-GCN+ADG\cite{cheng2020eccv} & 90.8 & 96.6\\
			PA-ResGCN-B19 \cite{song2020stronger} & 90.9 & 96.0 \\
			DDGCN\cite{korban2020ddgcn} & 91.1 & \textbf{97.1}\\
			Dynamic GCN\cite{ye2020dynamic} & 91.5 & 96.0 \\
			MS-G3D\cite{liu2020disentangling} & 91.5 & 96.2\\
			\hline\hline
			\textbf{CTR-GCN} & \textbf{92.4} & 96.8\\
			\hline
		\end{tabular}
	\end{center}
	\vspace{-0.2cm}
	\caption{Classification accuracy comparison against state-of-the-art methods on the NTU RGB+D dataset.}
	\label{tab:ntu}
	%\vspace{-0.3cm}
\end{table}

\begin{table}
	\begin{center}
		\begin{tabular}{l c}
			\hline
			\multirow{2}{*}{\textbf{Methods}} & \textbf{Northwestern-UCLA} \\  &\textbf{Top-1 (\%)}\\
			\hline\hline
			Lie Group\cite{veeriah2015differential}  & 74.2\\
			Actionlet ensemble\cite{wang2013learning} & 76.0\\
			HBRNN-L\cite{du2015hierarchical} & 78.5\\
			Ensemble TS-LSTM\cite{lee2017ensemble} & 89.2\\
			\hline
			AGC-LSTM\cite{si2019attention} & 93.3\\
			Shift-GCN\cite{cheng2020skeleton} & 94.6\\
			DC-GCN+ADG\cite{cheng2020eccv} & 95.3 \\
			\hline\hline
			\textbf{CTR-GCN} & \textbf{96.5}\\
			\hline
		\end{tabular}
	\end{center}
	\vspace{-0.2cm}
	\caption{Classification accuracy comparison against state-of-the-art methods on the Northwestern-UCLA dataset.}
	\label{tab:kinetics}
	\vspace{-0.5cm}
\end{table}

Many state-of-the-art methods employ a multi-stream fusion framework. We adopt same framework as \cite{cheng2020skeleton,ye2020dynamic} for fair comparison. Specifically, we fuse results of four modalities, \ie, joint, bone, joint motion, and bone motion.

We compare our models with the state-of-the-art methods on NTU RGB+D 120, NTU RGB+D and NW-UCLA in Tables \ref{tab:ntu120}, \ref{tab:ntu} and $\ref{tab:kinetics}$ respectively. On three datasets, our method outperforms all existing methods under nearly all evaluation benchmarks. On NTU-RGB+D 120, our model with joint-bone fusion achieves state-of-the-art performance, and our CTR-GCN outperforms current state-of-the-art Dynamic GCN \cite{ye2020dynamic} by 1.6\% and 2.0\% on the two benchmarks respectively. Notably, our method is the first to model channel-wise topologies dynamically which is very effective in skeleton-based action recognition.

%Table \ref{tab:ntu120} and Table \ref{tab:ntu} compare non-graph methods and graph-based methods. Table \ref{tab:kinetics} compares graph-based methods with a single stream and multiple streams. On the NTU RGB+D, our models achieve comparable performances with state-of-the-art models with similar computation cost. Since the dataset is close to saturation, it's hard to surpass the state-of-the-art performance largely.

%On two larger datasets NTU RGB+D 120 and Kinetics Skeleton 400, our models exceed the current state-of-the-art methods significantly. It's worth noting that (1) on NTU RGB+D 120, 4s CTR-GCN-small surpasses 4s DC-GCN+ADG with only 17\% computation cost. (2) on NTU RGB+D 120, 2s CTR-GCN-small achieves comparable results with the well-designed, light-weight model 4s shift-GCN with half computation cost while 4s CTR-GCN-small exceeds 4s shift-GCN significantly with similar computation cost. (3) on Kinetics Skeleton 400, the single joint stream of CTR-GCN-big achieves comparable results with two-stream MS-G3D with only 36\% computation cost. DDGCN achieves 38.1\% but it utilizes other modal features (\ie, bone, motion and bone motion features) in the reasoning. In conclusion, our models outperform well-designed heavy or light models by simply adjusting the model size, showing the superiority of our method. Notably, our method is the first to model channel-wise topologies in a refinement approach which has been proven to be effective in skeleton-based action recognition.

\section{Conclusion}
In this work, we present a novel channel-wise topology refinement graph convolution (CTR-GC) for skeleton-based action recognition. CTR-GC learns channel-wise topologies in a refinement way which shows powerful correlation modeling capability. Both mathematical analysis and experimental results demonstrate that CTR-GC has stronger representation capability than other graph convolutions. On three datasets, the proposed CTR-GCN outperforms state-of-the-art methods.

\noindent\textbf{Acknowledgment} This work is supported by the National Key R\&D Plan (No.2018YFC0823003), Beijing Natural Science Foundation (No.L182058), the Natural Science Foundation of China (No.61972397,62036011,61721004), the Key Research Program of Frontier Sciences, CAS (No.QYZDJ-SSW-JSC040),  National Natural Science Foundation of China (No.U2033210).

{\small
\bibliographystyle{ieee_fullname}
\bibliography{egbib}
}

\clearpage
\section*{Supplemental Materials for Channel-wise Topology Refinement Graph Convolution for Skeleton-Based Action Recognition}  % **** Enter the paper title here

\thispagestyle{empty}

%%%%%%%%% BODY TEXT - ENTER YOUR RESPONSE BELOW
This supplemental materials include details about formula derivation, architecture setting, more visualizations and other ablation studies. Specifically, we give the derivation from Equation {\color{blue}12} to {\color{blue}13} and from Equation {\color{blue}8} to {\color{blue}14}. Then we show the detailed architecture of CTR-GCN, including input size, output size and specific hyperparameters of each block. Moreover, we visualize shared topologies and channel-specific correlations. At last, we conduct ablation studies on the effect of CTR-GC's number per block, temporal convolution, and analyze the performance of different graph convolutions on hard classes.

\section*{Formula Derivation}

We first give the derivation from Equation {\color{blue}12} to {\color{blue}13}. The Equation {\color{blue}12} is
\vspace{-0.16cm}
\begin{equation}
\vspace{-0.16cm}
\mathbf{z^k_i}=\sum_{v_j \in \mathcal{N}(v_i)}\mathbf{p_{ij}}\odot (\mathbf{x^k_jW}),
\end{equation}
where $\mathbf{z^k_i} \in \mathbb{R}^{1\times C'}$ is the output feature of $v_i$ and $\mathbf{p_{ij}}\in \mathbb{R}^{1\times C'}$ is the channel-wise relationship between $v_i$ and $v_j$. $\mathbf{x^k_j}\in \mathbb{R}^{1\times C}$ is the input feature of $v_j$ and $\mathbf{W}\in \mathbb{R}^{C\times C'}$ is weight matrix. The $c$-th element of $\mathbf{z^k_i}$ is formulated as
\vspace{-0.16cm}
\begin{align}
z^k_{ic}&=\sum_{v_j \in \mathcal{N}(v_i)}p_{ijc}\mathbf{(x^k_jW)_c}=\sum_{v_j \in \mathcal{N}(v_i)}p_{ijc}\mathbf{(x^k_jw_{:,c})}\notag\\
\vspace{-0.16cm}
&=\sum_{v_j \in \mathcal{N}(v_i)}\mathbf{x^k_j}(p_{ijc}\mathbf{w_{:,c}}),
\vspace{-0.16cm}
\label{equ:static}
\end{align}
where $p_{ijc}$ is the $c$-th element of $\mathbf{p_{ij}}$.  $\mathbf{(x^k_jW)_c} \in \mathbb{R}^1$ is the $c$-th element of $\mathbf{x^k_jW}$ and $\mathbf{w_{:,c}}\in \mathbb{R}^{C\times 1}$ is the $c$-th column of $\mathbf{W}$. Therefore, $\mathbf{z^k_i}$ can be formulated as
\vspace{-0.16cm}
\begin{align}
\mathbf{z^k_i}&=\left[
\begin{array}{c}
\sum_{v_j \in \mathcal{N}(v_i)}\mathbf{x^k_j}(p_{ij1}\mathbf{w_{:,1}})\\
\vdots\\
\sum_{v_j \in \mathcal{N}(v_i)}\mathbf{x^k_j}(p_{ijC'}\mathbf{w_{:,C'}})
\end{array}
\right]^T \notag\\
&=\sum_{v_j \in \mathcal{N}(v_i)}\mathbf{x^k_j}([p_{ij1}\mathbf{w_{:,1}},\cdots, p_{ijC'}\mathbf{w_{:,C'}}]),
\label{equ:static_reform}
\end{align}
which is the same as Equation {\color{blue}13}.

Then we give the derivation from Equation {\color{blue}8} to {\color{blue}14}. We add sample index $\mathbf{k}$ in Equation {\color{blue}8}, which is formulated as
\vspace{-0.16cm}
\begin{equation}
\vspace{-0.16cm}
\mathbf{Z^k}=[\mathbf{R^k_1\tilde{x}^k_{:,1}} || \mathbf{R^k_2\tilde{x}^k_{:,2}}||\cdots||\mathbf{R^k_{C'}\tilde{x}^k_{:,C'}}].
\end{equation}

The $c$-th column of $\mathbf{Z^k}\in \mathbb{R}^{N\times C'}$ can be formulated as
\vspace{-0.16cm}
\begin{equation}
\vspace{-0.16cm}
\mathbf{z^k_{:,c}}=\mathbf{R^k_{c}\tilde{x}^k_{:,c}}=\mathbf{R^k_c}\mathbf{(X^kW)_{:,c}}=\mathbf{R^k_c}\mathbf{(X^kw_{:,c})},
\end{equation}
where $\mathbf{X^k}\in \mathbb{R}^{N\times C}$ is the input feature. The $i$-th element of $\mathbf{z^k_{:,c}}$, \ie, the $c$-th element of $v_i$'s output feature is
\begin{align}
z^k_{ic}&=\mathbf{r^k_{i,:,c}}\mathbf{(X^kw_{:,c})}=\sum_{v_j \in \mathcal{N}(v_i)}r^k_{ijc}\mathbf{(x^k_jw_{:,c})} \notag\\
&=\sum_{v_j \in \mathcal{N}(v_i)}\mathbf{x^k_j}(r^k_{ijc}\mathbf{w_{:,c}}),
\label{equ:ctrgc}
\end{align}
where $\mathbf{r^k_{i,:,c}}\in \mathbb{R}^{1\times N}$ is the $i$-th row of $\mathbf{R^k_c}\in \mathbb{R}^{N\times N}$. It can be seen that Equation \ref{equ:ctrgc} has the similar form with Equation \ref{equ:static}. Thus Equation \ref{equ:ctrgc} can be reformulated to the similar form with Equation \ref{equ:static_reform}, which is
\vspace{-0.16cm}
\begin{equation}
\vspace{-0.16cm}
\mathbf{z^k_i}=\sum_{v_j \in \mathcal{N}(v_i)}\mathbf{x^k_j}([r^k_{ij1}\mathbf{w_{:,1}},\cdots, r^k_{ijC'}\mathbf{w_{:,C'}}]).
\label{equ:ctrgc_reform}
\end{equation}

It can be seen that Equation \ref{equ:ctrgc_reform} is the same as Equation {\color{blue}14}, \ie, Equation {\color{blue}8} can be reformulated to Equation {\color{blue}14}.

\begin{table}[!h]
	\begin{center}
		\begin{tabular}{c|c|c}
			\hline
			Layers & Output Sizes & Hyperparameters \\
			\hline
			Basic Block 1 & M$\times$T$\times$N & $\begin{bmatrix}\text{SM: 3, C} \\ \text{TM: C, C, 1}\end{bmatrix}$\\
			\hline
			Basic Block 2& M$\times$T$\times$N & $\begin{bmatrix}\text{SM: C, C} \\ \text{ TM: C, C, 1}\end{bmatrix}$\\
			\hline
			Basic Block 3& M$\times$T$\times$N & $\begin{bmatrix}\text{SM: C, C} \\ \text{ TM: C, C, 1}\end{bmatrix}$\\
			\hline
			Basic Block 4& M$\times$T$\times$N & $\rm\begin{bmatrix}\text{SM: C, C} \\ \text{TM: C, C, 1}\end{bmatrix}$\\
			\hline
			Basic Block 5& M$\times$$\frac{\text{T}}{\text{2}}\times$N & $\begin{bmatrix}\text{SM: C, 2C}\\ \text{TM: 2C, 2C, 2}\end{bmatrix}$\\
			\hline
			Basic Block 6& M$\times$$\frac{\text{T}}{\text{2}}\times$N & $\begin{bmatrix}\text{SM: 2C, 2C }\\ \text{TM: 2C, 2C, 1}\end{bmatrix}$\\
			\hline
			Basic Block 7& M$\times$$\frac{\text{T}}{\text{2}}\times$N & $\begin{bmatrix}\text{SM: 2C, 2C }\\ \text{TM: 2C, 2C, 1}\end{bmatrix}$\\
			\hline
			Basic Block 8& M$\times$$\frac{\text{T}}{\text{4}}\times$N & $\begin{bmatrix}\text{SM: 2C, 4C} \\ \text{TM: 4C, 4C, 2}\end{bmatrix}$\\
			\hline
			Basic Block 9& M$\times$$\frac{\text{T}}{\text{4}}\times$N & $\begin{bmatrix}\text{SM: 4C, 4C }\\ \text{TM: 4C, 4C, 1}\end{bmatrix}$\\
			\hline
			Basic Block 10& M$\times$$\frac{\text{T}}{\text{4}}\times$N & $\begin{bmatrix}\text{SM: 4C, 4C}\\ \text{TM: 4C, 4C, 1}\end{bmatrix}$\\
			\hline
			Classification& 1$\times$1$\times$1 & $ \begin{bmatrix}\text{global averge pool }\\ \text{$n_c$-d fc} \\ \text{softmax}\end{bmatrix}$\\
			\hline
		\end{tabular}
	\end{center}
	%\vspace{-0.3cm}
	\caption{Detailed architecture of CTR-GCN. M, T, and N refer to the number of people, the length, and the number of joints of input sequences. ``SM" and ``TM" indicate the spatial modeling module and temporal modeling module respectively. The two numbers after SM are the input channel and output channel of SM. The three numbers after TM are the input channel, output channel and temporal stride. $n_c$ is the number of action classes.}
	\label{tab:archi}
	%\vspace{-0.3cm}
\end{table}

\section*{Detailed Architecture}

The detailed architecture of the proposed CTR-GCN is shown in Table \ref{tab:archi}, CTR-GCN contains ten basic blocks and a classification layer which consists of a global average pooling, a fully connected layer and a softmax operation. M refers to the number of people in the sequences, which is set to 2, 2, and 1 for NTU RGB+D, NTU RGB+D 120, and NW-UCLA respectively. In a sequence, M skeleton sequences are processed independently by ten basic blocks and are average pooled by the classification layer to obtain the final score. T and N refer to the length and the number of joints of input skeleton sequences, which are \{64, 25\}, \{64, 25\} and \{52, 20\} for NTU-RGB+D, NTU-RGB+D 120, and NW-UCLA respectively. C is the basic channel number which is set to 64 for CTR-GCN. ``SM" and ``TM" indicate the spatial modeling module and temporal modeling module respectively. The two numbers after SM are the input channel and output channel of SM. The three numbers after TM are the input channel, output channel and temporal stride. At the Basic Blocks 5 and 8, the strides of convolutions in temporal modeling module (TM) are set to 2 to reduce the temporal dimension by half. $n_c$ is the number of action classes, which is 60, 120, 10 for NTU-RGB+D, NTU-RGB+D120, and NW-UCLA respectively. 

\section*{Visualization}
\begin{figure}[t]
	\centering
	\includegraphics[width=\columnwidth]{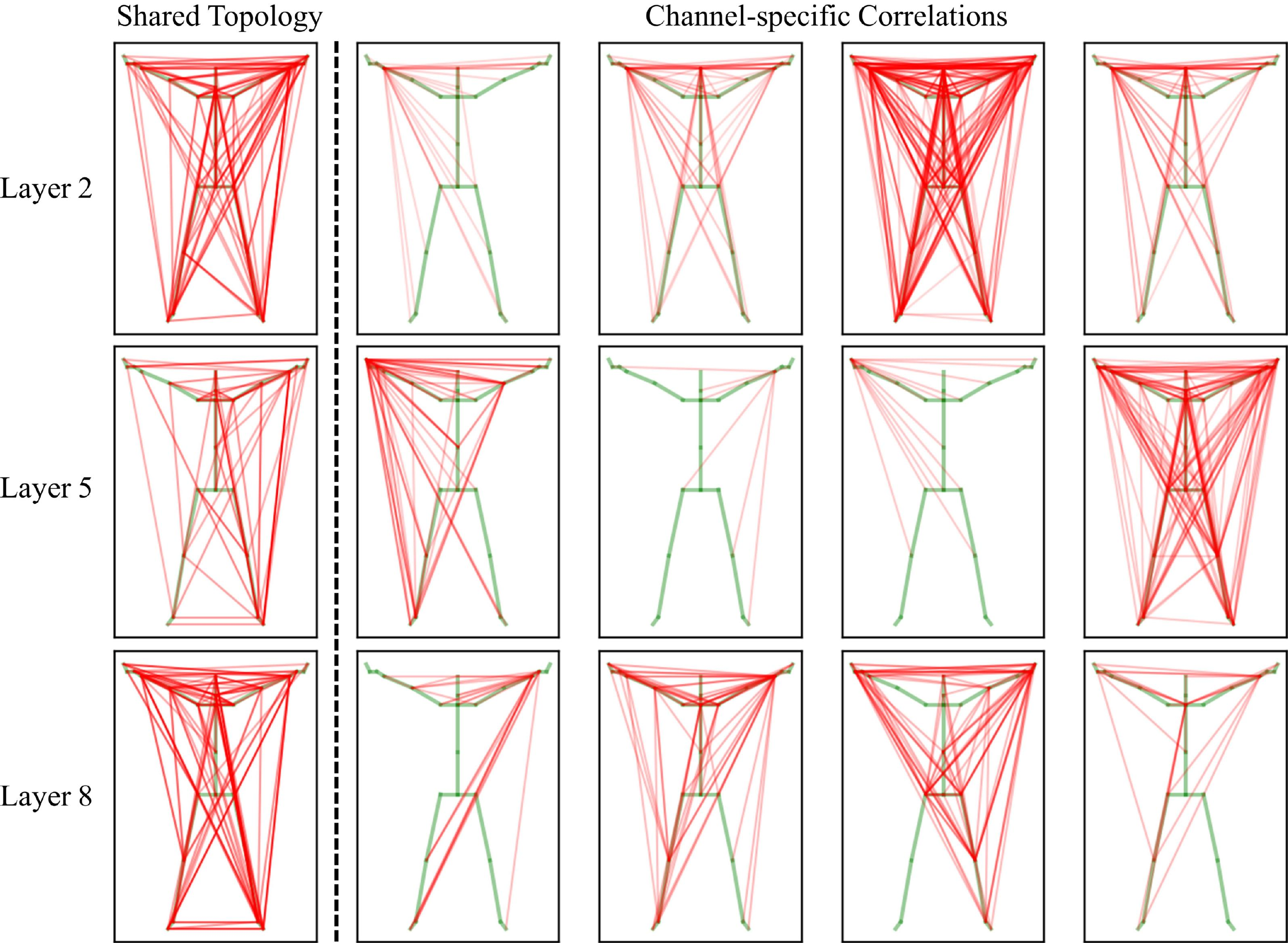} % Reduce the figure size so that it is slightly narrower than the column. Don't use precise values for figure width.This setup will avoid overfull boxes.
	\caption{Visualization of the shared topologies and channel-specific correlations. The green lines show the natural connections of human skeleton. The intensity of red lines indicates the connection strength of correlations.}
	\label{fig:visualize}
\end{figure}

As shown in Figure \ref{fig:visualize}, we visualize the shared topologies and channel-specific correlations of our CTR-GCN. The input sample belongs to ``typing on a keyboard". It can be seen that (1) the shared topologies in three layers tend to be coarse and dense, which captures global features for recognizing actions; (2) the channel-specific correlations varies with different channels, indicating that our CTR-GCN models individual joints relationships under different types of motion features; (3) most channel-specific correlations focus on two hands, which capture subtle interactions on hands and are helpful for recognizing ``typing on a keyboard".

\section*{Ablation Study}
\begin{table}
	\begin{center}
		\begin{tabular}{c c l}
			\hline
			\textbf{Number} & \textbf{Param.} & \ \ \textbf{Acc (\%)} \\
			\hline\hline
			3(CTR-GCN) & 1.46M & \ \ \ 84.9 \\
			\hline
			1 & 0.85M & \ \ \ 84.3 $^{\downarrow 0.5}$\\
			2 & 1.15M & \ \ \ 84.7 $^{\downarrow 0.2}$\\
			4 & 1.76M & \ \ \ 85.2 $^{\uparrow 0.3}$ \\
			5 & 2.07M & \ \ \ \textbf{85.4} $^{\mathbf{\uparrow 0.5}}$\\
			6 & 2.37M & \ \ \ 85.0 $^{\uparrow 0.1}$ \\
			\hline
		\end{tabular}
	\end{center}
	\caption{Comparisons of model performances with different number of CTR-GCs.}
	\label{tab:num_ctrgcs}
\end{table}

\noindent \textbf{Effect of CTR-GC's number.} In CTR-GCN, we use three CTR-GCs for fair comparison with other methods (e.g., AGCN, MSG3D), which mostly use three or more GCs to increase model capacity. To verify the effectiveness of CTR-GC's number to our method, We test the model with 1~6 CTR-GCs. As shown in Table \ref{tab:num_ctrgcs}, accuracies first increase due to increased model capacity, but drops at 6 CTR-GCs, which may be caused by overfitting.

\begin{table}
	\begin{center}
		\begin{tabular}{c l}
			\hline
			\textbf{Temporal Modeling} & \ \ \textbf{Acc (\%)} \\
			\hline\hline
			Temporal Conv(CTR-GCN) & \ \ \ 84.9 \\
			\hline
			Temporal Pooling & \ \ \ 72.8 $^{\downarrow 12.1}$\\
			\hline
		\end{tabular}
	\end{center}
	\caption{Comparisons of model performances with different number of CTR-GCs.}
	\label{tab:temp_conv}
\end{table}

\noindent \textbf{Effect of temporal convolutions.} It's a common practice to use (multi-scale) temporal convolutions for temporal modeling in skeleton-based action recognition. To validate the effect of temporal convolutions, we try to use global average pooling for temporal modeling. As shown in Table \ref{tab:temp_conv}, the performance drops from 84.9\% to 72.8\%, probably because the pooling loses too much temporal information to extract joints' trajectory features effectively.

\begin{figure}[t]
	\centering
	\includegraphics[width=\columnwidth]{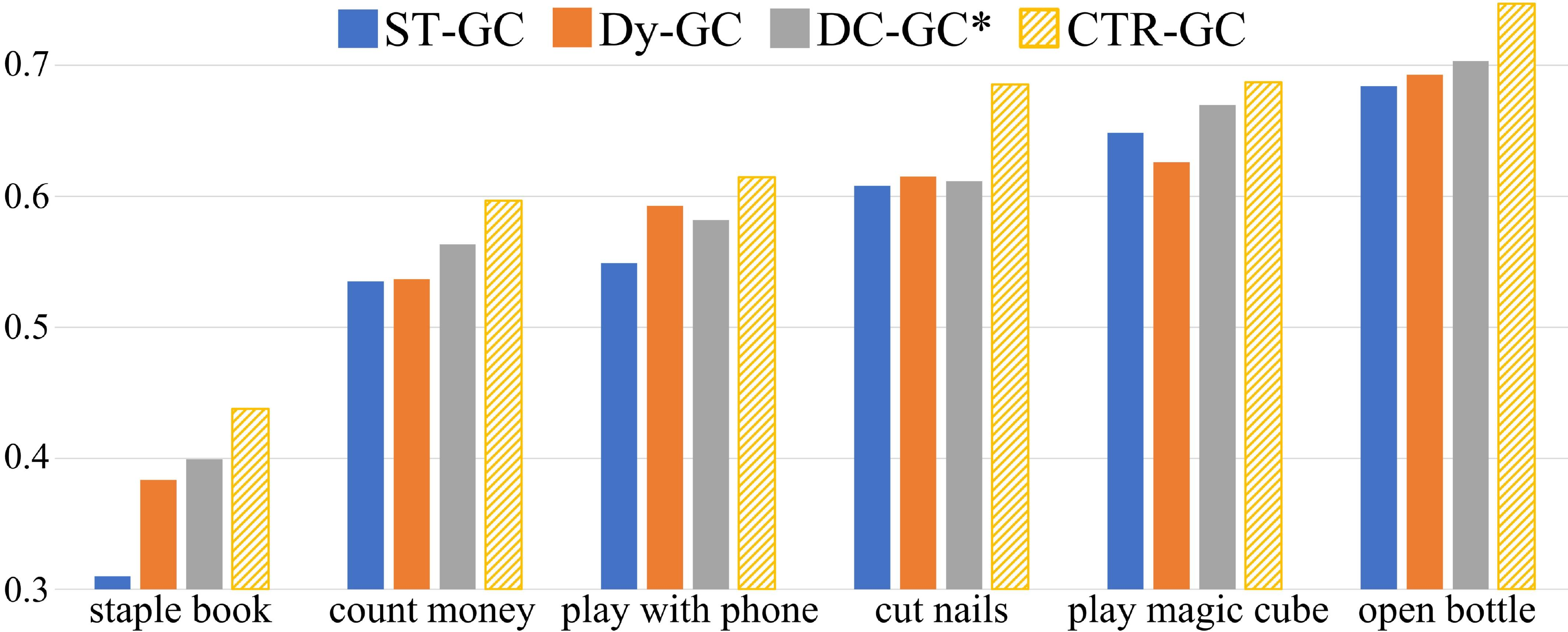} % Reduce the figure size so that it is slightly narrower than the column. Don't use precise values for figure width.This setup will avoid overfull boxes.
	\caption{Comparison of classification accuracy of different graph convolutions on hard action classes.} 
	\vspace{-0.3cm}
	\label{fig:acc}
\end{figure}

\noindent \textbf{Performance on hard classes.} We further analyze the performance of different graph convolutions on hard classes on NTU-RGB+D 120, \ie, ``staple book", ``count money", ``play with phone" and ``cut nails", ``playing magic cube" and ``open bottle". These actions mainly involve subtle interactions between fingers, making them difficult to be recognized correctly. As shown in Figure \ref{fig:acc}, CTR-GC outperforms other graph convolutions on all classes. Especially, CTR-GC exceeds other methods at least by 7.03\% and 4.36\% on ``cut nails" and ``open bottle" respectively, showing that, compared with other GCs, our CTR-GC can effectively extract features of subtle interactions and classify them more accurately.
\end{document}